\documentclass[runningheads]{llncs}
\usepackage[T1]{fontenc}
\usepackage{graphicx}

\usepackage{amsmath}
\usepackage{amssymb}
\usepackage{algorithm}
\usepackage{algorithmic}

\usepackage{booktabs}
\usepackage{multirow}
\usepackage{makecell}
\usepackage{siunitx}
\usepackage{subcaption}
\usepackage{hyperref}

\usepackage{bbding}
\usepackage{color}

\begin{document}
\title{LiSA: Leveraging Link Recommender to Attack Graph Neural Networks via Subgraph Injection}
\titlerunning{LiSA: \underline{Li}nk Recommender-\underline{S}ubgraph Injection \underline{A}ttack}
%

\author{Wenlun Zhang\inst{1}\orcidID{0000-0003-1570-082X} \and
Enyan Dai\inst{2}\Envelope\orcidID{0000-0001-9715-0280} \and
Kentaro Yoshioka\inst{1}\orcidID{0000-0001-5640-2250}}

\authorrunning{W. Zhang et al.}

%

\institute{Keio University, Kanagawa, Japan \\ \email{\{wenlun\_zhang,kyoshioka47\}@keio.jp} \and
Hong Kong University of Science and Technology (Guangzhou), Guangzhou, China \\ \email{enyandai@hkust-gz.edu.cn}}

\maketitle              
\begin{abstract}
Graph Neural Networks (GNNs) have demonstrated remarkable proficiency in modeling data with graph structures, yet recent research reveals their susceptibility to adversarial attacks. Traditional attack methodologies, which rely on manipulating the original graph or adding links to artificially created nodes, often prove impractical in real-world settings. This paper introduces a novel adversarial scenario involving the injection of an isolated subgraph to deceive both the link recommender and the node classifier within a GNN system. Specifically, the link recommender is mislead to propose links between targeted victim nodes and the subgraph, encouraging users to unintentionally establish connections and that would degrade the node classification accuracy, thereby facilitating a successful attack. To address this, we present the LiSA framework, which employs a dual surrogate model and bi-level optimization to simultaneously meet two adversarial objectives. Extensive experiments on real-world datasets demonstrate the effectiveness of our method. \textbf{Code is available at} \href{https://github.com/Wenlun-Zhang/LiSA}{\textit{https://github.com/Wenlun-Zhang/LiSA}}.

\end{abstract}
\section{Introduction}

In recent years, Graph Neural Networks (GNNs) have shown remarkable proficiency in modeling graph-structured data. GNNs primarily operate on a message-passing mechanism. This process of iterative information aggregation enables GNNs to effectively update node representations while incorporating the structural properties. However, the powerful message-passing mechanism of GNNs also brings its share of challenges and limitations. Malicious attackers can intentionally perturb the graph structure and features. Such deliberately tampered information can propagate among nodes, leading to a degraded performance of GNNs~\cite{RL-S2V,Nettack,Meta_Attack,FGA,AFGSM,GF-Attack}. Recognizing the importance of designing trustworthy GNNs, investigating various attack methods has become an essential initial undertaking~\cite{GNN_Attack_Review}. Consequently, there is a growing focus on adversarial attacks targeting GNNs within academic and industrial communities. Among these, manipulation attacks~\cite{Nettack} have been demonstrated to be particularly effective. These attacks involve adding or deleting edges, or modifying node features in the original graph, all within a constrained budget.

Nonetheless, adversaries frequently encounter practical hurdles when attempting to manipulate these data, as it usually requires direct access to the backend data, a task that is not easily achievable. On the other hand, node injection attacks (NIAs)~\cite{NIPA,G-NIA,TDGIA} present a more feasible approach. Such attacks degrade classification performance by introducing malicious nodes linked to target nodes, spreading harmful attributes. Within citation networks, adversaries may craft counterfeit papers and insert references to target papers, ultimately causing them to be misclassified. This kind of emerging attack method have demonstrated high success rates and are increasingly seen as a more practical method in real-world scenarios. Injecting fake nodes is often more feasible than modifying data on existing nodes, making these techniques particularly insidious.

\begin{figure}[ht]
\centering
\includegraphics[width=\textwidth]{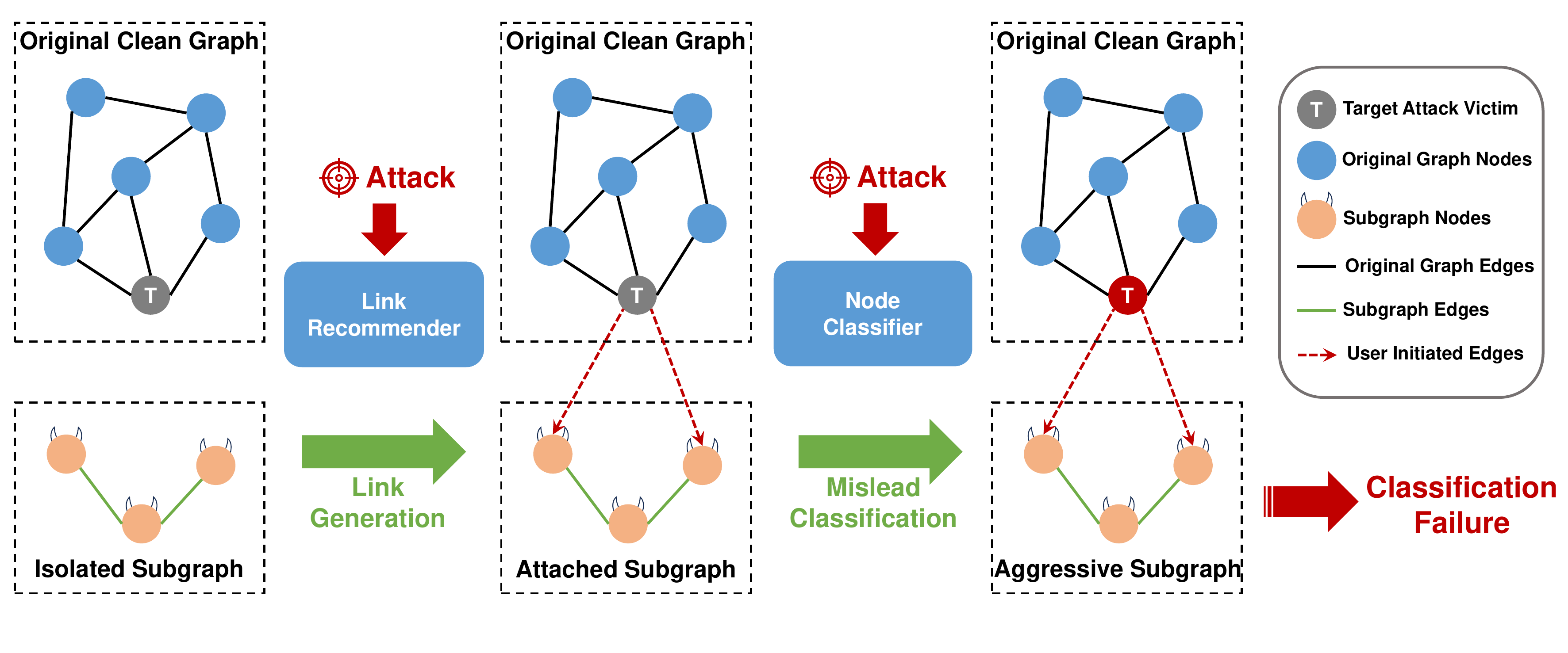}
\caption{SIA scenario via link recommender.}
\label{Fig_SIA}
\end{figure}

There is no doubt about the practicality and effectiveness of NIAs. Nevertheless, current research often focuses on ideal conditions, which implies that adversaries can freely establish connections between target nodes and injected fake nodes. However, this assumption does not universally hold across various applications. Especially in applications where security is critical, like social networks, fake users who connect with the target user can subtly alter the ground truth label of the target user, leading to targeted advertising. However, it is not certain that target users will approve connection request with fake accounts. Thus, evaluating these attack models under the assumption of cost-free linking can be misleading, potentially leading to an overestimation of the attack outcomes. These concerns highlighted above have motivated us to focus on enhancing the likelihood of adversaries successfully linking fake nodes to the target nodes. Given the prevalence of link recommendation algorithms within GNN-based applications, we explore the potential to exploit these link recommenders to facilitate malicious link formation. The proposed subgraph injection attack (SIA) scenario is shown in Fig. \ref{Fig_SIA}. Our method does not manually add edges between target victim nodes and the aggressor fake nodes. Instead, it deceives the link recommender of GNN into actively initiating links formation. Once these links are established, the classification performance on the target victim nodes suffers a considerable decline, ultimately enabling a successful attack. This novel approach to adversarial attacks is crucial for real-world applications yet remains largely unexplored. Moreover, the uncertain feasibility of attacking multiple GNN models simultaneously poses a significant challenge in solving this problem.

In this work, we investigate a unique SIA scenario, which injects an isolated subgraph to deceive both the link recommender and the node classifier in a GNN simultaneously. Specifically, it aims to mislead the link recommender into creating adversarial links from targeted victim nodes to the subgraph, thereby undermining the accuracy of node classification. Our contributions in this study are outlined as follows: \textbf{(1)} We explore an innovative attack scenario utilizing the link recommender to compromise the node classification accuracy within a GNN. \textbf{(2)} We introduce \underline{Li}nk Recommender-\underline{S}ubgraph Injection \underline{A}ttack (LiSA), a pioneering framework crafted to effectively generate subgraphs by employing a dual surrogate model and bi-level optimization. \textbf{(3)} Through rigorous experimentation on diverse real-world datasets, we validate the effectiveness and applicability of our proposed methodologies.

\section{Related Works}

Adversarial attacks on GNNs aim to exploit model vulnerabilities, resulting in errors or inaccurate predictions. These attacks are generally categorized into evasion, poisoning, and backdoor attacks, each characterized by distinct methodologies and consequences. Evasion attacks occur during the test phase, where adversaries manipulate the input graph data to lead the GNN to make incorrect predictions~\cite{Nettack,TDGIA,MinMax}. These manipulations could involve strategic alterations in the graph structure or node features, significantly affecting the performance of the GNN. The stealth of evasion attacks, which do not require changes to the trained model, makes them particularly challenging to detect. Following evasion attacks, poisoning attacks are conducted during the training phase of the GNN. In this scenario, attackers inject the training set with crafted data~\cite{RL-S2V,NIPA}, either by adding, deleting edges, or modifying nodes. These alterations subtly distort the training process, leading the model to unwittingly integrate these adversarial modifications. The performance of a model consequently deteriorates, introducing specific errors or biases, undermining its reliability and accuracy. Lastly, backdoor attacks are implemented by inserting hidden triggers in the GNN during its training phase~\cite{Backdoor_Attack_1,Backdoor_Attack_2,UGBA}. When activated in new data, these triggers cause the model to generate incorrect results, while maintaining normal functionality under standard conditions. This makes backdoor attacks particularly deceptive.

Unfortunately, none of the existing attack methods can fully demonstrate their capabilities in practical scenarios. Attackers do not have the capability to modify the original graph in evasion and poisoning attacks. Even with the utilization of relatively accessible malicious injection techniques, the failure to generate links immediately undermines the attack, leading to a considerable degradation in their efficacy. Therefore, we propose leveraging the link recommender to facilitate the generation of links between the target victim and injected nodes. The involvement of link recommender enhances the likelihood of establishing connections between the aggressor subgraph and the target victim nodes, consequently leading to a higher success rate in practical attacks.

\section{Preliminaries}


\subsection{Notations}

We define the original graph as $G_o = (A_o, F_o)$, with adjacency matrix $A_o \in \mathbb{R}^{N_o \times N_o}$ and feature matrix $F_o \in \mathbb{R}^{N_o \times D}$. The target victim nodes, $v_t \in V$, are selected for the attack. An isolated subgraph, denoted as $G_s = (A_s, F_s)$, is injected, forming the poisoned graph $G_p = (A_p, F_p)$: 

\begin{equation}
A_p = \left [ \begin{array}{cc} A_o & 0 \\ 0 & A_s \\ \end{array} \right ], \quad F_p = \left [ \begin{array}{cc} F_o \\ F_s \\ \end{array} \right ]
\label{Eq_Poisoned_Graph}
\end{equation}

\noindent where $A_p \in \mathbb{R}^{(N_o + N_s) \times (N_o + N_s)}$ and $F_p \in \mathbb{R}^{(N_o + N_s) \times D}$ is the adjacency matrix and features of the poisoned graph.

\subsection{Threat Model and Problem Formulation}
\textbf{Attacker's Goal.} The objective of the adversary is to deceive both the link recommender, conceptualized as a form of link prediction model, and the node classifier simultaneously in a GNN system. This is achieved through the strategic injection of a deliberately crafted subraph, which will mislead the link prediction model to form connections with the target nodes, thereby influencing classification accuracy thereafter.
\newline
\textbf{Attacker's Knowledge and Capability.} Attackers lack specific knowledge regarding the GNN models employed, including the particular type of link prediction and node classification model utilized within the system. Nevertheless, attackers possess access to the original attributed graph as well as a restricted set of training and validation labels. Attackers are capable of introducing an isolated subgraph to the original graph. However, any alteration to the original graph or establishment of connections between the original graph and the subgraph is strictly prohibited. Additionally, the features and node degrees of the subgraph must closely resemble those of the original graph to maintain inconspicuousness.
\newline
\textbf{Problem.} We assume the presence of a link recommender $f_\phi$ in the GNN, utilizing a subset of edges from core users in the original graph to suggest connections. This results in an updated graph $G_r$, with:

\begin{equation}
G_r = (\hat{A}, F_p), \quad \hat{A} = f_{\phi}(G_p)
\label{Eq_Recommender}
\end{equation}

\noindent The SIA strategically generates adversarial edges between the target victim nodes and the isolated subgraph, updating the adjacency matrix to:

\begin{equation}
\hat{A} = \left[ \begin{array}{cc} A_o & A_l \\ {A}_{l}^{T} & A_s \\ \end{array} \right]
\label{Eq_Adversarial_Link_Generation}
\end{equation}

\noindent Here, $A_l \in \mathbb{R}^{N_s \times N_o}$ represents the edges generated via link recommender, and the rank of $A_l$ increases from zero, which indicates that some adversarial edges have been generated successfully. Finally, the node classification model $f_{\theta}$ provides predictions based on the updated graph:

\begin{equation}
\hat{y_{v_t}} = f_{\theta}(G_r)
\label{Eq_Node_Classification}
\end{equation}

\noindent where $\hat{y_{v_t}}$ is the predicted class on target victim nodes. The ultimate objective of SIA is to compromise classification accuracy, making $\hat{y_{v_t}}$ diverge from the ground truth labels.

\subsection{Preliminary Analysis}

At the outset of validating the proposed SIA scenario, we initiate its characterization employing a heuristic approach called GraphCopy. The core concept of GraphCopy focuses on creating a subgraph through a process that duplicates target nodes and their environmental surroundings, under the assumption that the message-passing mechanisms within GNNs would produce highly similar representations for these replicated structures:

\begin{equation}
G_s = \text{Copy}(\mathcal{N}_{n}(v_t)), \quad \mathcal{N}_{n}(v_t) = \{u \in V | \text{dist}(u, v_t) \leq n\}
\label{Eq_GraphCopy}
\end{equation}

\noindent Here, $\text{Copy}()$ is the replicating operation, $\mathcal{N}_{n}(v_t)$ represent the n-hop neighborhoods set on $v_t$, and $\text{dist}(u, v_t)$ gives the minimum distance between nodes $u$ and $v_t$. Such similarity seemingly facilitates the formation of links between the original graph and subgraph. However, simply generating links does not achieve our primary goal of reducing the accuracy of node classification. To address this, we introduce a perturbation in the form of Gaussian noise to the node features of the duplicated subgraph. By fine-tuning the number of hops and the intensity of perturbation, we can achieve a balance that facilitates link generation while reducing node classification accuracy, thereby achieving the adversarial objectives. Although we verified that GraphCopy can partially achieve the adversarial goal of SIA, it faces two significant limitations that reduce its effectiveness and practicality in real-world scenarios. First, simply duplicating the graph and altering node features does not ensure a high success rate for adversarial attacks. Second, replicating the n-hop neighborhood of a target node leads to a subgraph size that increases quadratically with the node degree. For nodes with high degrees, the attack budget increase substantially, making GraphCopy impractical for use in dense or large-scale networks. To maximize the effectiveness of attacks while minimizing costs, there is a strong need for an optimization-driven method to refine the subgraph for SIA.

\section{Methodology}

\begin{figure}[ht]
\includegraphics[width=\textwidth]{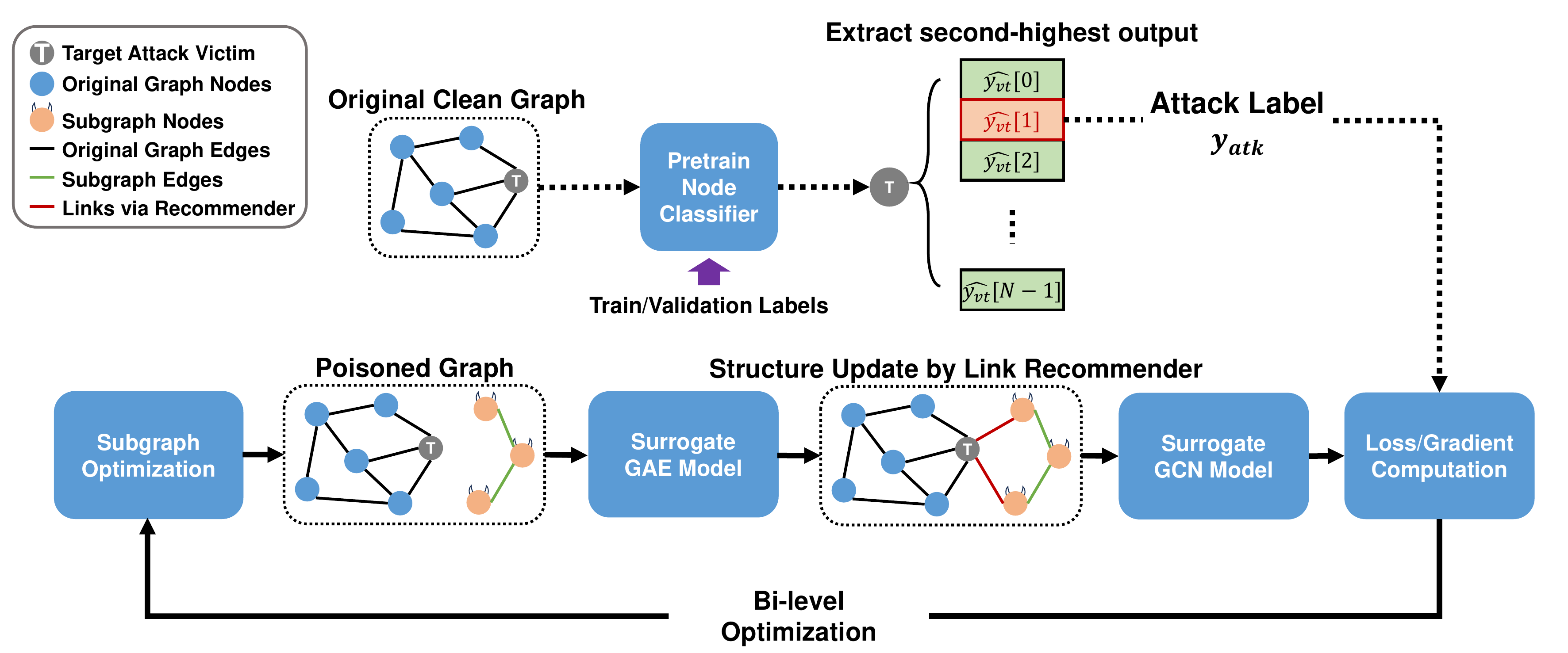}
\caption{An overview of LiSA framework.}
\label{Fig_LiSA}
\end{figure}

In this section, we investigate an optimization-based approach for generating subgraphs. In the SIA scenario, achieving dual adversarial goals within the SIA scenario presents significant technical challenges. Intuitively, generating links is more likely when there is a close resemblance in the node representations between the target victim nodes and the nodes of the aggressor subgraph, whereas a greater disparity in these representations increases the chances of the victim nodes being misclassified. As a result, a more deliberate subgraph optimization technique is essential to balance two conflicting objectives and identify the optimal subgraph for SIA. To this end, we introduce the LiSA framework, as depicted in Fig. \ref{Fig_LiSA}.

\subsection{LiSA Framework}

The LiSA framework is designed to challenge the performance of two GNN models through a strategic deployment of dual-surrogate models for bi-level optimization in a black-box attack context. We select the GAE~\cite{VGAE} and GCN~\cite{GCN} as the surrogate models for link prediction and node classification, aiming to refine the adjacency matrix and feature attributes of the subgraph to meet two critical optimization goals concurrently: \textbf{(i)} the generation of adversarial edges between the original graph and the subgraph via the link recommender, and \textbf{(ii)} negatively affect node classification performance on target victim nodes once they are connected to the subgraph. To navigate these objectives, the attack loss function of LiSA includes two terms, balanced by a factor $\alpha$, to regulate the contribution of each model:

\begin{equation}
\mathcal{L}_{atk} = \mathcal{L}^{cls}_{atk}(G_r; \theta) + \alpha \mathcal{L}^{link}_{atk}(G_p; \phi)
\label{Eq_Attack_Loss}
\end{equation}

\noindent The loss related to link prediction, $\mathcal{L}^{link}_{atk}$, is defined as a reconstruction loss with Q negative sampling, aiming to mislead the link predictor into establishing adversarial links $a_{ij}$ between the original graph and the subgraph:

\begin{equation}
\mathcal{L}^{link}_{atk} = \mathcal{L}_{recon} = - \sum_{i \in V} \sum_{j \in V} \log \hat{a}_{ij} - \sum_{n=1}^{Q} \cdot \mathrm{E}_{v_{n} \sim P_n(v_i)} \log (1 - \hat{{a_n}}_{ij})
\label{Eq_Attack_Link_Loss}
\end{equation}

\noindent For the node classification, the attack loss, $\mathcal{L}^{cls}_{atk}$, seeks to reduce the chance of the target nodes being correctly classified by employing a cross-entropy loss with the attack label $y_{atk}$:

\begin{equation}
\mathcal{L}^{cls}_{atk} = - \sum_{v_t} y_{atk} \cdot \log(\hat{y_{v_t}})
\label{Eq_Attack_Class_Loss}
\end{equation}

\noindent Notably, the approach begins by pretraining a node classifier using a limited label set, subsequently swapping the top-2 output classes to define the attack labels~\cite{Flip_Label}. This approach circumvents the traditional use of negative cross-entropy loss used in GradMax, which complicates the task of balancing the two objectives. The continuously increasing adversarial loss for classification, $\mathcal{L}^{cls}_{atk}$, usually fails to maintain balance between these two loss terms. By giving precedence to the second-highest output class for the attack, we foster a more stable training environment. The comprehensive training strategy is outlined as follows:

\begin{equation}
\begin{aligned}
\min_{A_s, F_s} \quad & \mathcal{L}^{cls}_{atk}(G_r; \theta^*) + \alpha \mathcal{L}^{link}_{atk}(G_p; \phi^*) \\
\text{s.t.} \quad & \theta^{*} = \mathop{\arg \min}_{\theta} \mathcal{L}^{cls}_{train}(f_{\theta}(G_r)), \quad & \phi^* = \mathop{\arg \min}_{\phi} \mathcal{L}^{link}_{train} (f_{\phi}(G_p)).
\end{aligned}
\label{Eq_Bi-Level}
\end{equation}

\noindent The method involves the simultaneous bi-level optimization of the two surrogate models, with optimizing space limited to the subgraph. 

\subsection{Training Algorithm}

The training procedure for the subgraph involves a two-tiered process with inner and outer iterations. In the inner iterations, we simultaneously train surrogate models for link prediction and node classification to optimize model parameters, while the outer iterations optimize the subgraph by alternately updating its adjacency matrix and node features. The inherent complexity of graph-structured data means that changes to the edges can significantly impact node representation. This alternating update methodology is necessary because it helps the training to achieve convergence. 

\subsubsection{Feature Optimization}

The optimization process for refining the features within the subgraph employs a conventional approach that utilizes the principles of gradient descent through backpropagation. The goal is to adjust the node features in a manner that directly contributes to the two adversarial objectives. Each iteration of feature optimization is guided by the following rule:

\begin{equation}
F_s = F_s - lr \cdot \nabla_{F_s} \mathcal{L}_{atk}
\label{Eq_Update_Feat}
\end{equation}

\noindent Here, $lr$ stands for the learning rate, and the gradient of the attack loss with respect to the subgraph features, $\nabla_{F_s} \mathcal{L}_{atk}$, directs the update, ensuring that each adjustment supports the two adversarial goals. Adjusting the balance factor $\alpha$ in the attack loss is crucial for aligning the feature updates with the dual aims of the attack. 

\subsubsection{Structure Optimization}

Optimizing the structure of the subgraph presents a more complex challenge compared to feature adjustments. Upon the initialization of subgraph, the total number of edges within it is fixed. To maintain this constant edge count during each iteration of structure optimization, we ensure that for every edge removed, a corresponding edge is added. This process is designed to preserve the integrity of the subgraph structure while aligning with our adversarial objectives. The approach to directing structural updates involves calculating a combined gradient of the adjacency matrix at the outset of each optimization cycle:

\begin{equation}
\nabla_{str} = \nabla_{A_p} \mathcal{L}_{atk}^{link} + \beta \cdot \nabla_{\hat{A}} \mathcal{L}_{atk}^{cls}
\label{Eq_Combined_Grad}
\end{equation}

\noindent In this formula, $\nabla_{A_p} \mathcal{L}_{atk}^{link}$ denotes the gradient of attack loss by the link prediction model with respect to the subgraph adjacency matrix $A_p$ before link recommender intervention, while $\nabla_{\hat{A}} \mathcal{L}_{atk}^{cls}$ represents the gradient of attack loss by the node classification model with respect to the post-update adjacency matrix $\hat{A}$ modified by the link recommender. The parameter $\beta$ acts as a balancing factor in the optimization of the structure. The structural update process involves ranking the combined gradient $\nabla_{str}$, then strategically adding and removing edges based on this ranking. Specifically, we add edges in locations where they are currently absent, prioritizing those with the lowest combined gradient values. Simultaneously, we remove edges from areas with the highest combined gradient values. 

\section{Experiments}

\subsection{Experiment Settings}

To evaluate the effectiveness of our attack methods, we perform experiments on five distinct datasets. These datasets include Cora and PubMed, both citation networks, Amazon-Photo and Computers, co-purchase networks from a popular e-commerce platform, and FacebookPagePage, representing a social network graph. We conduct an experimental verification of our LiSA framework using GAE and VGAE~\cite{VGAE} for link recommender, along with GCN~\cite{GCN}, SGC~\cite{SGC}, and GraphSAGE~\cite{GraphSAGE} as node classifiers. Our attack methodology is applied across all datasets with the presumption that the target user would establish links with nodes of top three linking scores in the entire poisoned graph. To assess the feasibility of our methods, we randomly select 1000 nodes from each original graph as target victim nodes. We then generate subgraphs and execute attacks on these nodes individually. This approach allows us to compute both the link success rate (LSR) and the overall attack success rate (ASR). The LSR is defined as the proportion where at least one adversarial edge is generated between the target node and subgraph, while the ASR refers to the misclassfication rate on these victim nodes. For the link recommender, we use 85\% of the edges for training, with half serving as positive edge labels with negative sampling for both subgraph optimization and attack verification. For node classification, 5\% of the nodes are randomly chosen for training labels. These labels are utilized in subgraph optimization and the pretraining of a node classifier to identify the second highest output class as the attack label. To maintain a consistent basis for comparison across different datasets, we employ a subgraph comprising five nodes for each target victim node attack. However, we tailor hyperparameters such as the number of subgraph edges $n_E$, and the balance parameters $\alpha$ and $\beta$ for each dataset, ensuring optimal performance and adaptability to the specific characteristics of each dataset. We also compare the LiSA framework with baseline of GraphCopy in the SIA scenario, and an additional baseline of conventional NIA scenario. This particular version of NIA incorporates the probability of successful link creation, aiming to demonstrate the influence of link generation on the success of NIAs. 

\begin{figure}[htbp]
    \centering
    \includegraphics[width=\textwidth]{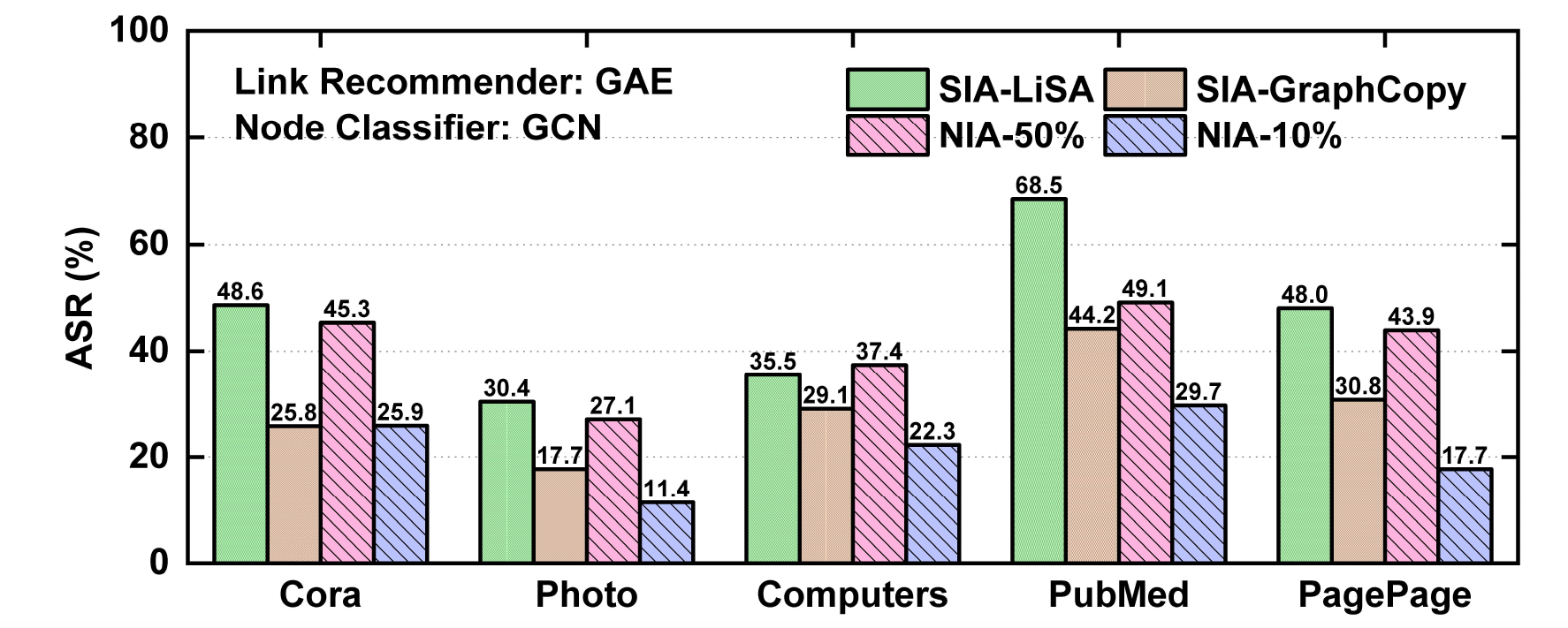}
    \caption{Attack outcome on GCN with SIA and NIA scenario. For SIA, the link recommender model is GAE.}
    \label{Fig_Attack_Outcome}
\end{figure}

\subsection{Attack Results}

The attack results using the LiSA framework are presented in Table \ref{Table_LiSA}, while the misclassification rates for the original clean graphs are also provided, serving as a reference for evaluating the impact of our attacks. The outcomes of the attack indicate that the subgraphs produced by the LiSA framework successfully establish links between target nodes and subgraphs, leading to a notable decrease in the accuracy of node classification performance. This demonstrates the effectiveness of the LiSA framework within the SIA scenario. As illustrated in Fig. \ref{Fig_Attack_Outcome}, the LiSA framework surpasses GraphCopy by achieving a higher ASR, thereby demonstrating its superiority over GraphCopy as an optimization-based method. Notably, the average attack budget for GraphCopy is proportional to the square of the average node degree in the dataset. In denser graphs, such as Amazon-Photo and Computers, attacking a single node requires a subgraph size exceeding 10\% of the original graph. Conversely, LiSA achieves higher ASR with a much smaller budget, requiring only five nodes for each target node attack, highlighting the efficiency and practicality in real-world scenarios.

Moreover, it is critical to compare the SIA strategy with the traditional NIA approach. While NIA is highly effective under the idealized assumption of a 100\% LSR, the ASR significantly decreases with the introduction of a variable LSR. When comparing the LiSA framework against the NIA, it becomes evident that the ASR of LiSA is comparable to that of NIA at a 50\% LSR, and it largely surpasses the performance of NIA at a 10\% LSR. This indicates that LiSA demonstrates superior attack efficacy in scenarios where successful link generation is a limiting factor of the attack. Consequently, in such instances, transitioning from the conventional NIA to the SIA strategy via the LiSA framework could be a wiser decision.

\begin{table*}[htbp]
\centering
\caption{Attack results (LSR \% | ASR \%) and clean graph misclassification (\%).}
\label{Table_LiSA}
\begin{tabular}{@{}ccccccc@{}}
\Xhline{1.5pt}
 & \textbf{Cora} & \textbf{Photo} & \textbf{Computers} & \textbf{PubMed} & \textbf{PagePage} \\
\Xhline{1.5pt}
\textbf{GCN} & 18.7 & 7.7 & 14.7 & 21.6 & 11.7 \\
\textbf{SGC} & 18.7 & 50.1 & 44.0 & 25.3 & 26.8 \\
\textbf{GraphSAGE} & 21.4 & 7.4 & 13.8 & 22.5 & 15.1 \\
\Xhline{1.5pt}
\textbf{GAE \& GCN} & 52.3 | 48.6 & 31.0 | 30.4 & 34.0 | 35.5 & 82.3 | 68.5 & 50.4 | 48.0 \\
\textbf{GAE \& SGC} & 52.1 | 56.1 & 34.7 | 55.5 & 37.7 | 43.5 & 79.9 | 68.8 & 51.6 | 61.6 \\
\textbf{GAE \& GraphSAGE} & 52.6 | 55.1 & 34.5 | 29.2 & 35.8 | 32.6 & 80.2 | 64.3 & 49.6 | 47.0 \\
\hline
\textbf{VGAE \& GCN} & 44.6 | 50.3 & 25.2 | 29.5 & 32.9 | 37.2 & 77.3 | 66.8 & 42.4 | 45.0 \\
\textbf{VGAE \& SGC} & 45.9 | 59.9 & 26.2 | 58.2 & 33.9 | 43.1 & 79.5 | 68.2 & 35.8 | 52.6 \\
\textbf{VGAE \& GraphSAGE} & 46.8 | 55.5 & 26.8 | 21.8 & 35.6 | 32.5 & 79.4 | 67.4 & 40.2 | 37.4 \\
\Xhline{1.5pt}
\end{tabular}
\end{table*}

\subsection{Ablation Studies}

\begin{figure}[htbp]
    \centering
    \begin{subfigure}[b]{0.49\columnwidth}
        \centering
        \includegraphics[width=\textwidth]{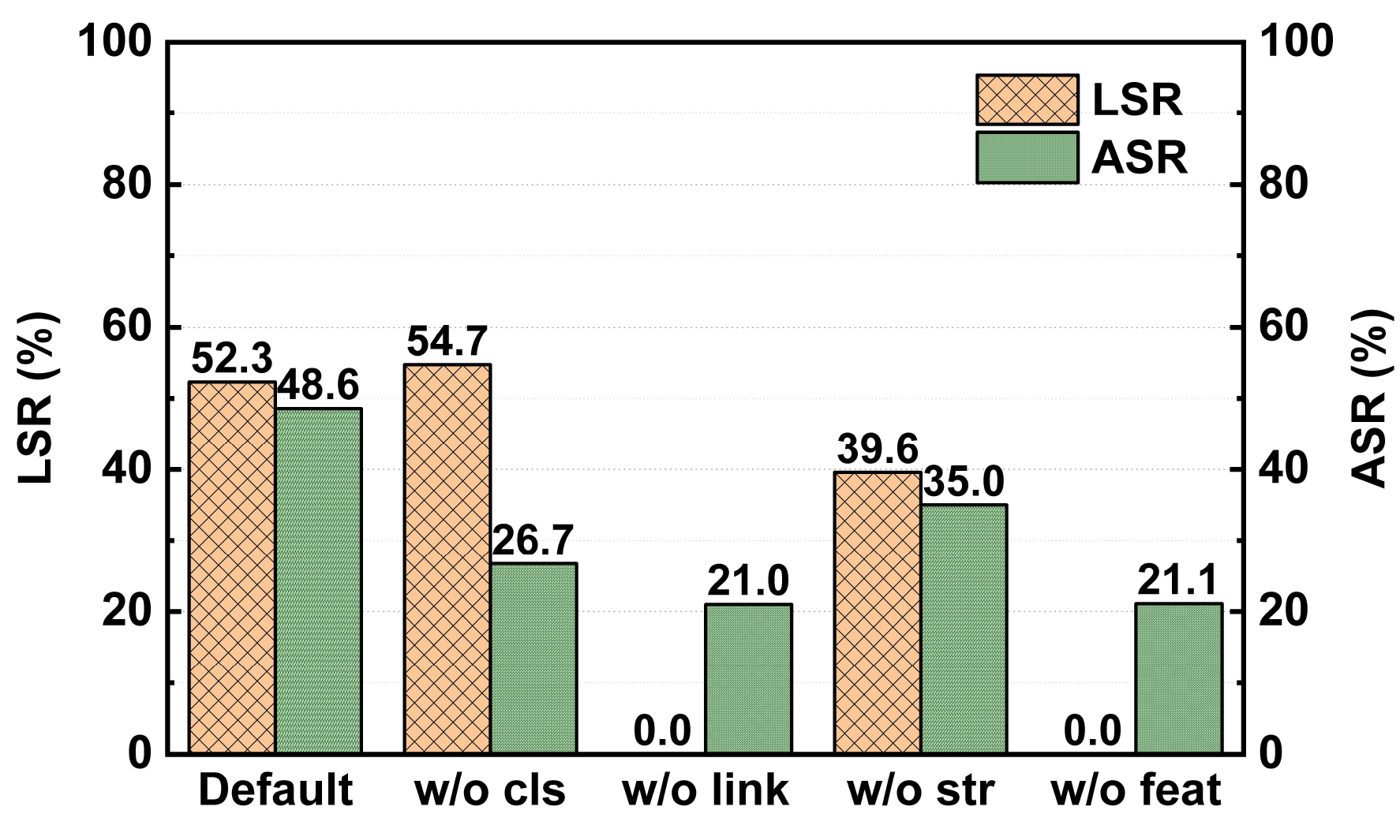}
        \caption{Cora}
        \label{Fig_Ablation_Cora}
    \end{subfigure}
    \hfill
    \begin{subfigure}[b]{0.49\columnwidth}
        \centering
        \includegraphics[width=\textwidth]{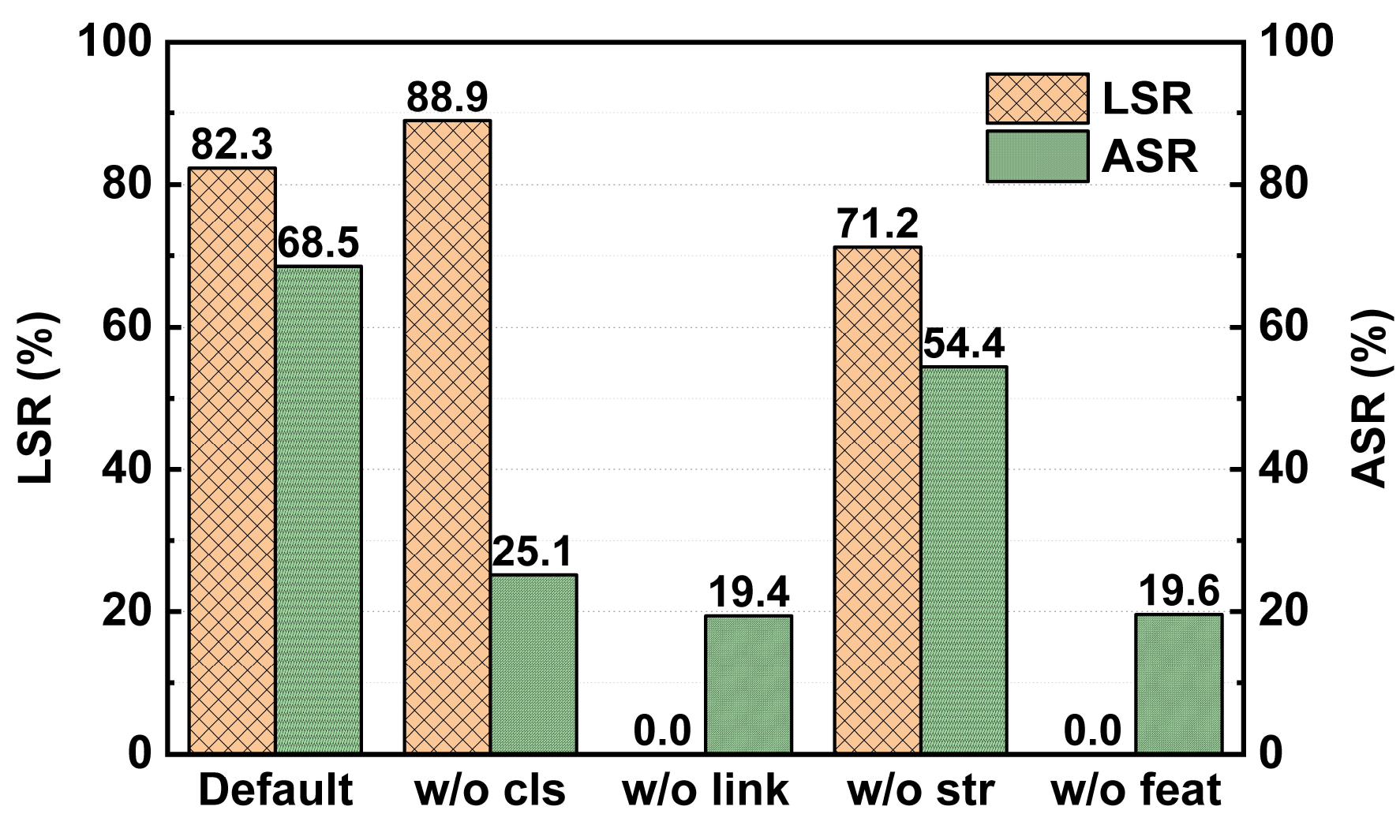}
        \caption{PubMed}
        \label{Fig_Ablation_PubMed}
    \end{subfigure}
    \caption{Ablation studies.}
    \label{Fig_Ablation}
\end{figure}

We perform ablation studies of LiSA on the Cora and PubMed dataset as depicted in Fig. \ref{Fig_Ablation}. This analysis involves experiments with variations of the LiSA framework, including the omission of the surrogate models for node classification and link prediction (denoted as w/o cls and w/o link, respectively). Additionally, we explore the impact of focusing solely on node features optimization (labeled as w/o str) and solely on adjacency matrix optimization with randomly initialized node features (labeled as w/o feat) to discern the specific contributions of these optimization strategies.

These findings reveal that adversarial links can be formed effectively through the use of the link prediction surrogate model, with the subgraph thus generated showing a marginal ASR improvement against the clean graph. Conversely, relying only on the node classification surrogate model fails to generate any links between target node and the subgraph. This highlights the significance of utilizing dual surrogate models, which can greatly enhance the attack to accomplish both adversarial objectives. Regarding optimization strategies, we observe that focusing exclusively on the structural aspects without adjusting the node features has no improvement on the attack. However, optimizing only the node features partially achieve the adversarial objectives, indicating that feature optimization plays a primary role in the SIA scenario. Nevertheless, structural optimization emerges as a critical supporting factor, enhancing the overall efficacy of the attack. Thus, the experimental results suggest that each component of LiSA is crucial to improve the attack performance.

\section{Conclusions}

This study presents a new attack scenario, along with a novel attack framework that utilizes link recommender to compromise the node classification accuracy of GNNs through subgraph injection. Unlike conventional attacks that require direct manipulation of the original graph or the addition of links to artificially created nodes, LiSA executes an attack indirectly by injecting an isolated subgraph, thereby enhancing practicality in real-world scenarios. Our experiments across various real-world datasets demonstrate the effectiveness of LiSA, illustrating significant degradation in GNN performance. 

\begin{credits}
\subsubsection{\ackname} This research was supported in part by the JST CREST JPMJCR21D2, JSPS Kakenhi 23H00467, Futaba Foundation, Asahi Glass Foundation, Telecommunications Advancement Foundation, and the KLL Ph.D. Program Research Grant.
\end{credits}

%
%
%
%
\bibliographystyle{splncs04}
\bibliography{reference}

\clearpage
\appendix

\section{Training Algorithm}
\label{Appendix_Algorithm}

The training algorithm of LiSA is shown in Algorithm \ref{algorithm_LiSA}.

\begin{algorithm}[htbp]
\caption{LiSA Algorithm Procedure}
\label{algorithm_LiSA}
\begin{algorithmic}[1]
\REQUIRE Original graph $G_o$, number of nodes $n_V$ and edges $n_E$ in the subgraph.
\ENSURE Generated subgraph $G_s$ optimized for adversarial attack.
\STATE Pretrain a node classifier to get attack label $y_{atk}$.
\STATE Randomly generate subgraph $G_s$ with $n_V$ nodes and $n_E$ edges. Integrate $G_s$ into $G_o$ to form the poisoned graph $G_p$.
\STATE Initialize surrogate models of node classification $\theta$ and link prediction $\phi$.
\WHILE{the optimization has not yet converged}
\FOR{$t = 1, 2, ..., N$}
\STATE Optimize $\phi$ by updating its parameters to minimize $\nabla_{\phi} \mathcal{L}^{\phi}_{train}$.
\STATE Optimize $\theta$ by updating its parameters to minimize $\nabla_{\theta} \mathcal{L}^{\theta}_{train}$.
\ENDFOR
\IF{the current iteration is even}
\STATE Update the adjacency matrix $A_s$ of the subgraph to refine its structure.
\ELSIF{the current iteration is odd}
\STATE Update the feature matrix $F_s$ of the subgraph to adjust its node features.
\ENDIF
\ENDWHILE
\RETURN The optimized subgraph $G_s$ for adversarial purposes.
\end{algorithmic}
\end{algorithm}

\section{GraphCopy Settings and Attack Results}
\label{Appendix_GraphCopy}

GraphCopy introduces a fundamental approach for conducting adversarial attacks, achieving the adversarial objectives to a certain degree, as demonstrated in Table \ref{Table_GraphCopy}. However, it faces two significant limitations that reduce its effectiveness and practicality in real-world scenarios. First, simply duplicating the graph and altering node features does not ensure a high success rate for adversarial attacks. Second, replicating the n-hop neighborhood of a target node leads to a subgraph size that increases quadratically with the node degree. For nodes with high degrees, the attack budget increase substantially, making GraphCopy impractical for use in dense or large-scale networks. In the experimental configuration of GraphCopy, we create the subgraph by replicating the 2-hop neighborhoods surrounding the target nodes and introducing a 10\% Gaussian perturbation to the features of the duplicated nodes. 

\begin{table*}[htbp]
\centering
\caption{GraphCopy attack results (LSR \% | ASR \%) and clean graph misclassification rate (\%).}
\label{Table_GraphCopy}
\begin{tabular}{@{}ccccccc@{}}
\Xhline{1.5pt}
 & \textbf{Cora} & \textbf{Photo} & \textbf{Computers} & \textbf{PubMed} & \textbf{PagePage} \\
\Xhline{1.5pt}
\textbf{GCN} & 18.7 & 7.7 & 14.7 & 21.6 & 11.7 \\
\textbf{SGC} & 18.7 & 50.1 & 44.0 & 25.3 & 26.8 \\
\textbf{GraphSAGE} & 21.4 & 7.4 & 13.8 & 22.5 & 15.1 \\
\Xhline{1.5pt}
\textbf{GAE \& GCN} & 13.3 | 25.8 & 30.5 | 17.7 & 45.3 | 29.1 & 61.8 | 44.2 & 27.4 | 30.8 \\
\textbf{GAE \& SGC} & 14.0 | 35.5 & 30.8 | 56.0 & 44.4 | 46.3 & 62.6 | 51.7 & 29.0 | 56.0 \\
\textbf{GAE \& GraphSAGE} & 15.3 | 33.0 & 31.5 | 12.9 & 43.9 | 17.8 & 63.3 | 41.1 & 27.6 | 20.6 \\
\hline
\textbf{VGAE \& GCN} & 29.9 | 36.8 & 40.7 | 17.6 & 48.8 | 29.8 & 64.1 | 44.7 & 50.0 | 25.4 \\
\textbf{VGAE \& SGC} & 30.3 | 42.6 & 39.4 | 56.8 & 50.9 | 45.9 & 65.1 | 49.7 & 44.8 | 50.2 \\
\textbf{VGAE \& GraphSAGE} & 31.1 | 39.9 & 39.8 | 14.1 & 50.8 | 20.5 & 63.6 | 42.7 & 43.6 | 23.0 \\
\Xhline{1.5pt}
\end{tabular}
\end{table*}

\section{Dataset Statistics}
\label{Appendix_Dataset}

The datasets employed in this study are enumerated in Table \ref{Table_Datasets}.

\begin{table*}[htbp]
\centering
\caption{Dataset statistics.}
\label{Table_Datasets}
\begin{tabular}{@{}l@{\extracolsep{15pt}}ccccc@{}}
\Xhline{1.5pt}
\textbf{Dataset}            & \textbf{Nodes} & \textbf{Edges}  & \textbf{Average Degree} & \textbf{Features} & \textbf{Classes} \\ 
\Xhline{1.5pt}
\textbf{Cora}               & 2,708  & 10,556  & 3.90           & 1,433     & 7       \\
\textbf{Photo}              & 7,650  & 238,162 & 31.13          & 746      & 8       \\
\textbf{Computers}          & 13,752 & 491,722 & 35.76          & 767      & 10      \\
\textbf{PubMed}             & 19,717 & 88,648  & 4.50           & 500      & 3       \\
\textbf{PagePage}           & 22,470 & 342,004 & 15.22          & 128      & 4       \\
\Xhline{1.5pt}
\end{tabular}
\end{table*}

\section{NIA Settings, Results and Analysis}
\label{Appendix_NIA}

To evaluate our method in the context of the traditional NIA scenario, we introduce a modified baseline version of NIA that takes into account the LSR. In this variation, each targeted victim node is paired with a single aggressor node, with each connection being subject to a regulated probability of successful formation. The influence of link generation on the effectiveness of the NIA is assessed by modulating LSR of 100\%, 50\%, and 10\%. Results of these assessments are detailed in Table \ref{Table_NIA}.

With a 100\% LSR, aggressor nodes are guaranteed to link to their target victims, resulting in a significantly high ASR, even when only a single aggressor node is employed. Under these conditions, link generation is not a critical factor in the attack scenario, and NIA demonstrates superior performance over SIA. However, at a 50\% LSR, the ASR notably declines as expected, showcasing the advantages of SIA wherein its ASR becomes comparable to that of NIA. When the LSR is reduced to just 10\%, the effectiveness of NIA is greatly diminished, making SIA a more viable strategy for achieving adversarial objectives.

\begin{table*}[htbp]
\centering
\caption{Attack results of NIA (ASR \%) and clean graph misclassification rate (\%).}
\label{Table_NIA}
\begin{tabular}{@{}ccccccc@{}}
\Xhline{1.5pt}
 & & \textbf{Cora} & \textbf{Photo} & \textbf{Computers} & \textbf{PubMed} & \textbf{PagePage} \\
\Xhline{1.5pt}
\textbf{Clean} & \textbf{GCN} & 18.7 & 7.7 & 14.7 & 21.6 & 11.7 \\
\textbf{Clean} & \textbf{SGC} & 18.7 & 50.1 & 44.0 & 25.3 & 26.8 \\
\textbf{Clean} & \textbf{GraphSAGE} & 21.4 & 7.4 & 13.8 & 22.5 & 15.1 \\
\Xhline{1.5pt}
\textbf{100\%} & \textbf{GCN} & 61.2 & 39.8 & 52.7 & 67.5 & 71.4 \\
\textbf{100\%} & \textbf{SGC} & 75.8 & 63.4 & 70.1 & 69.0 & 74.4 \\
\textbf{100\%} & \textbf{GraphSAGE} & 38.6 & 11.0 & 16.4 & 26.0 & 17.0 \\
\hline
\textbf{50\%} & \textbf{GCN} & 45.3 & 27.1 & 37.4 & 49.1 & 43.9 \\
\textbf{50\%} & \textbf{SGC} & 53.5 & 60.3 & 55.1 & 50.4 & 46.0 \\
\textbf{50\%} & \textbf{GraphSAGE} & 35.0 & 9.2 & 15.1 & 25.3 & 15.8 \\
\hline
\textbf{10\%} & \textbf{GCN} & 25.9 & 11.4 & 22.3 & 29.7 & 17.7 \\
\textbf{10\%} & \textbf{SGC} & 26.4 & 56.6 & 51.0 & 30.8 & 28.5 \\
\textbf{10\%} & \textbf{GraphSAGE} & 24.5 & 8.1 & 14.1 & 23.8 & 15.4 \\
\Xhline{1.5pt}
\end{tabular}
\end{table*}

\section{Impacts of Subgraph Size and User Added Links}

\begin{figure}[htbp]
    \centering
    \begin{subfigure}[b]{0.49\columnwidth}
        \centering
        \includegraphics[width=\textwidth]{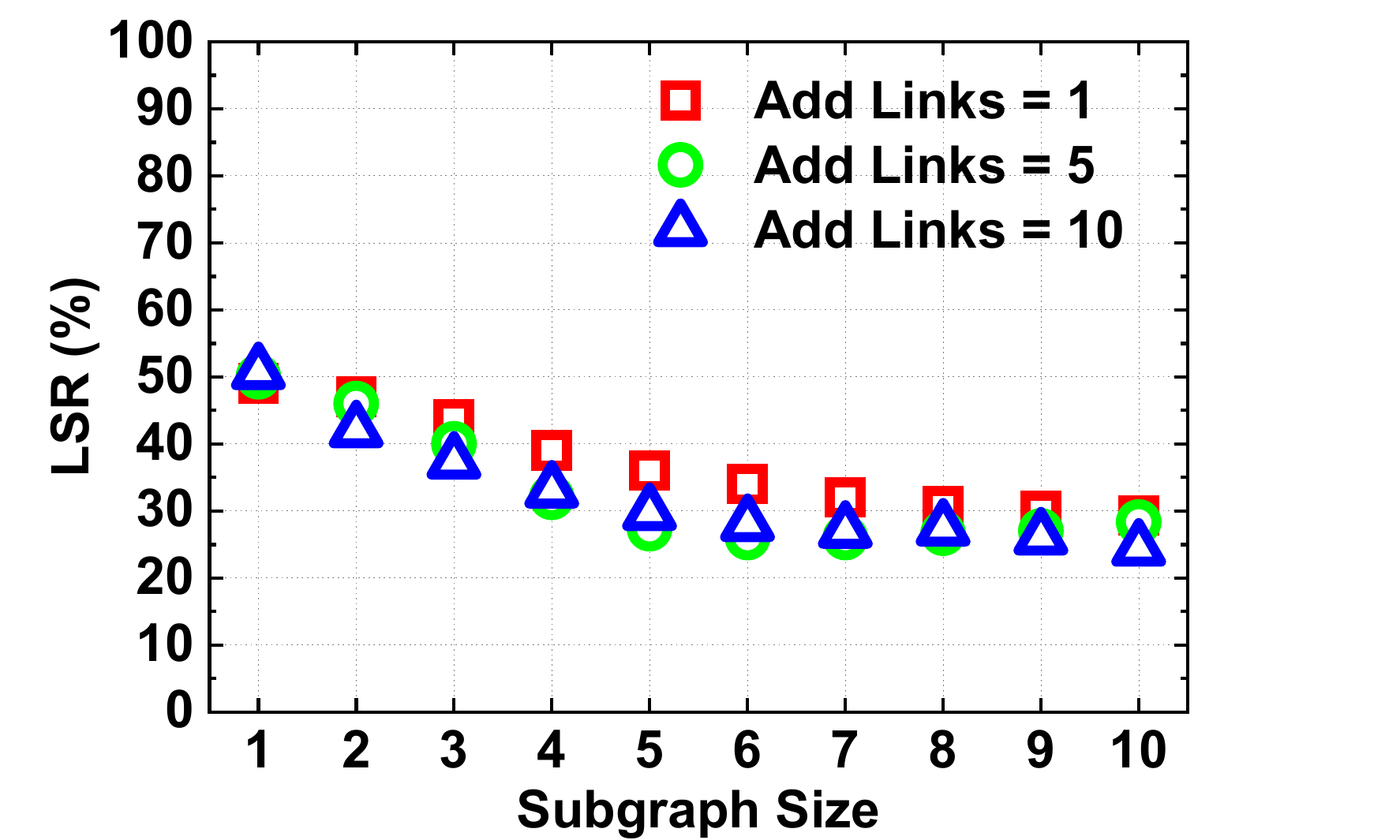}
        \caption{LSR (\%)}
        \label{Fig_LSR}
    \end{subfigure}
    \hfill
    \begin{subfigure}[b]{0.49\columnwidth}
        \centering
        \includegraphics[width=\textwidth]{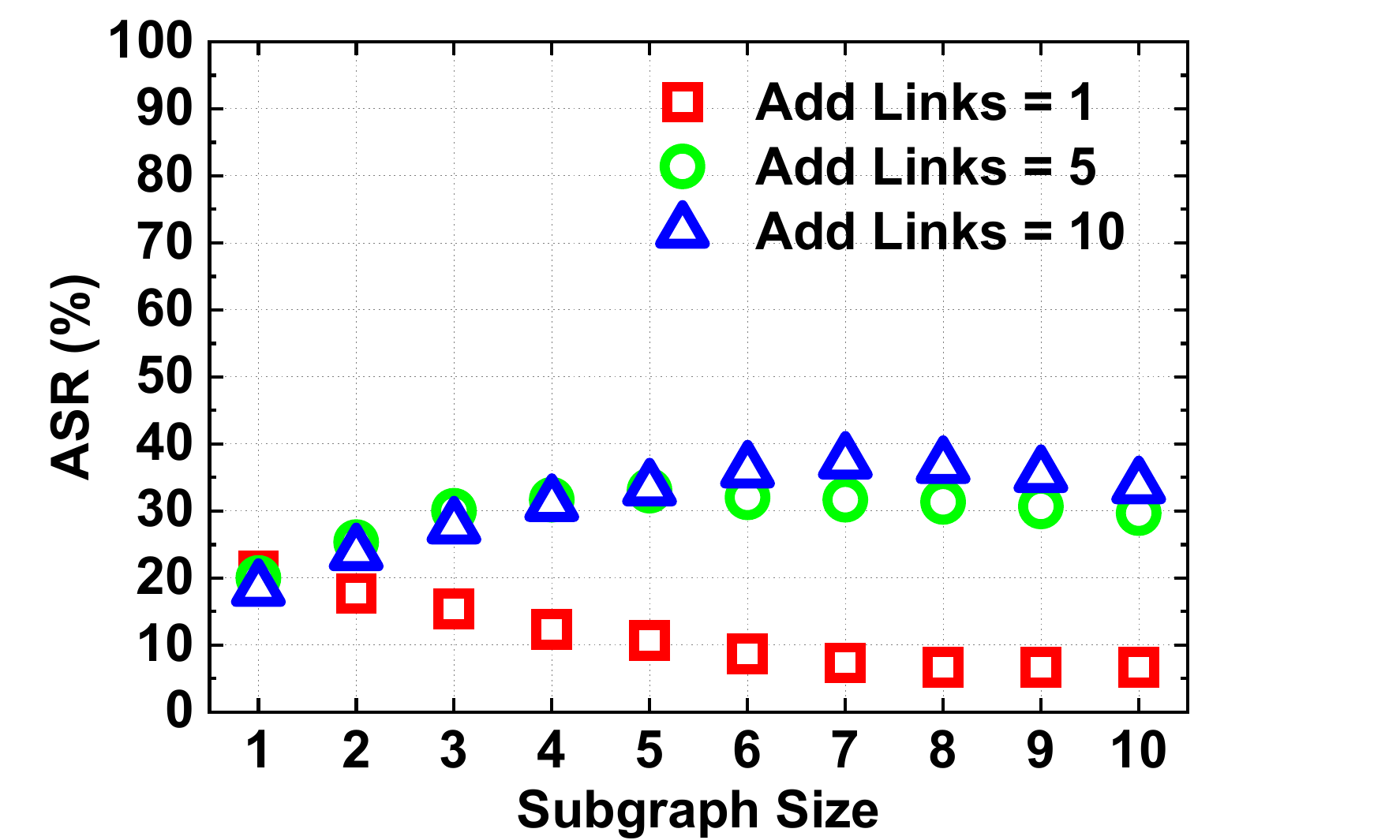}
        \caption{ASR (\%)}
        \label{Fig_ASR}
    \end{subfigure}
    \caption{Impacts of subgraph size and user added links on Amazon-Photo.}
    \label{Fig_Subgraph_Size}
\end{figure}

We conduct a series of experiments on the Amazon-Photo dataset to investigate how subgraph size impact the attack. Newly generated links are chosen based on top linking scores in the experiments. Our initial observation reveal a decline in LSR as the subgraph size increased, a trend depicted in Fig. \ref{Fig_Subgraph_Size}(a). This decline likely stems from the challenges associated with optimizing larger subgraphs, especially when adjusting both structure and features. However, we noted a stabilization in LSR when the subgraph size exceeds 5. Another interesting finding related to the response of ASR to the number of links added by users, as illustrated in Fig. \ref{Fig_Subgraph_Size}(b). With the addition of a single link, we observe that the ASR is highest when the subgraph consist of a single node, decreasing as the subgraph size expands, possibly due to the corresponding reduction in LSR. On the other hand, with five links added, the ASR improve as the subgraph grow to five nodes, before experiencing a slight decrease with further increases in subgraph size. This phenomenon can be attributed to the rise in aggressor nodes linked to the target, enhancing effectiveness of the attack. However, exceeding the subgraph size with available links prompts the link recommender to seek top-scoring nodes from the original graph for link formation, which strengthens the resilience of target node to the attack. Taking into account these effects, we found that the optimal ASR of 10 links was attained when the subgraph size was 7.

\section{Subgraph Analysis}

\begin{figure}[htbp]
    \centering
    \begin{subfigure}[b]{0.49\columnwidth}
        \centering
        \includegraphics[width=\textwidth]{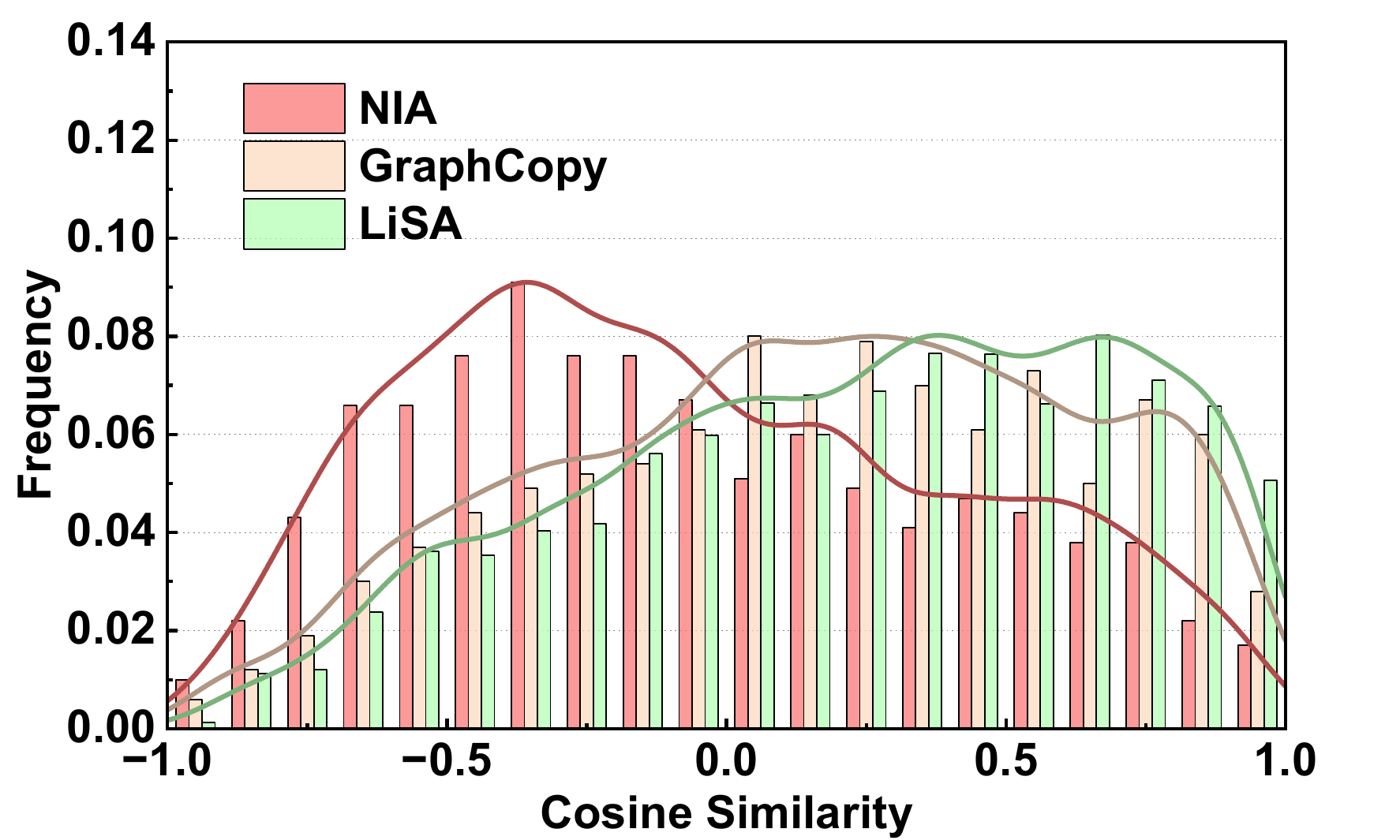}
        \caption{Cora}
        \label{Fig_Subgraph_Cora}
    \end{subfigure}
    \hfill
    \begin{subfigure}[b]{0.49\columnwidth}
        \centering
        \includegraphics[width=\textwidth]{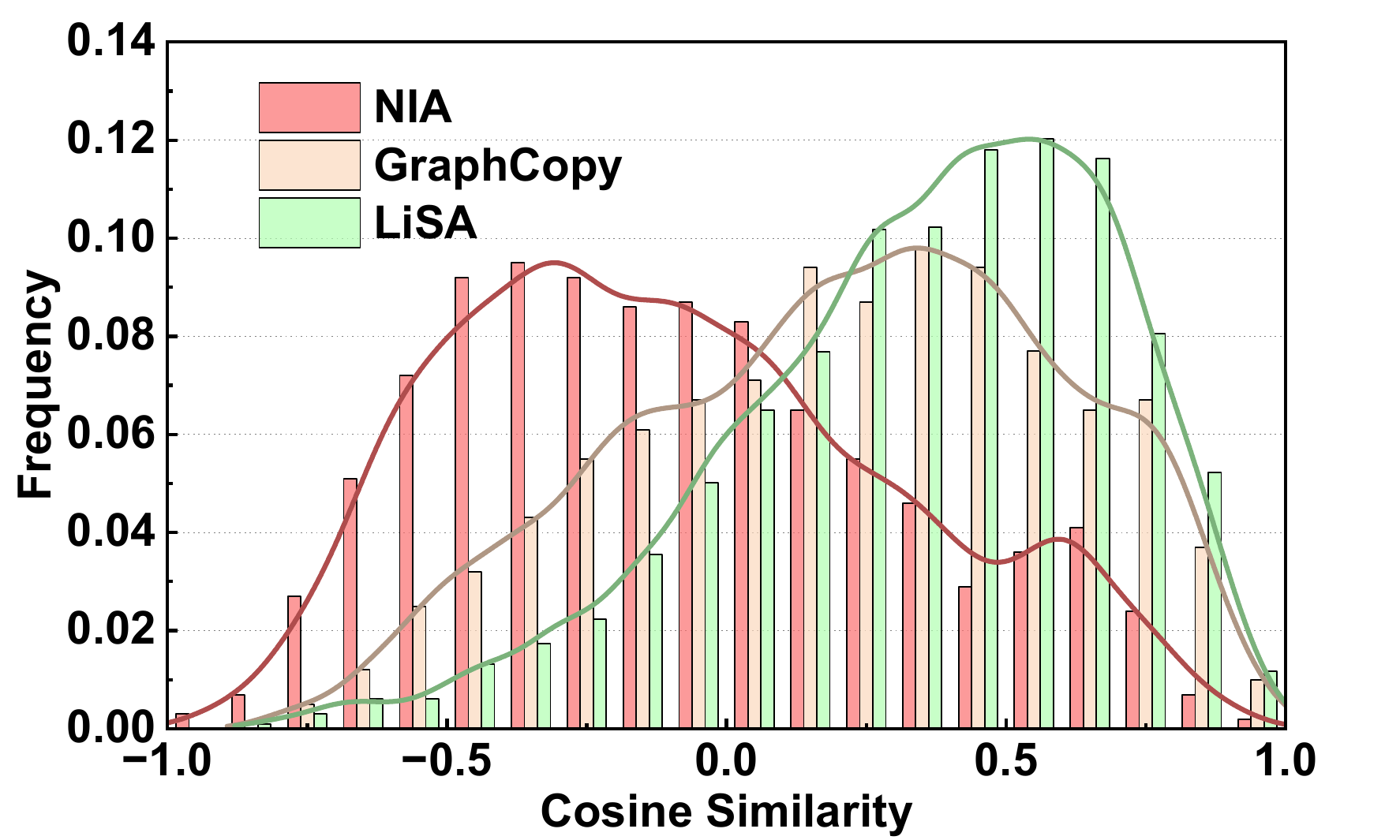}
        \caption{PubMed}
        \label{Fig_Subgraph_PubMed}
    \end{subfigure}
    \caption{Subgraph Analysis.}
    \label{Fig_Subgraph}
\end{figure}

To generate links between the target node and the subgraph, we expect the node embedding of the subgraph to be similar to that of the target victim node. To test this hypothesis, we train a GAE model on the original clean graph, use its GCN encoder to extract the node embedding of the target node and subgraph nodes, and calculate the cosine similarity between them. Experiments are conducted on both the Cora and PubMed datasets, and the similarity distributions are compared with those of NIA and GraphCopy. As shown in Fig. \ref{Fig_Subgraph}, nodes trained using NIA tend to have a negative similarity with the target nodes, with the distribution peaking at -0.4, indicating a low similarity and resulting in a very low LSR. For SIA, we observe that the node embedding tend to show high similarity, with the distribution peaks shifting to positive. Comparing LiSA with GraphCopy, we find that LiSA carefully refines the subgraph, further shifting the peak by approximately 0.4 compared to GraphCopy. As a result, the subgraph nodes with extremely high similarity are proposed to the target nodes by the link recommender, leading to a successful attack.

\section{Parameter Sensitivity Analysis}

\begin{figure}[htbp]
    \centering
    \begin{subfigure}[b]{0.49\columnwidth}
        \centering
        \includegraphics[width=\textwidth]{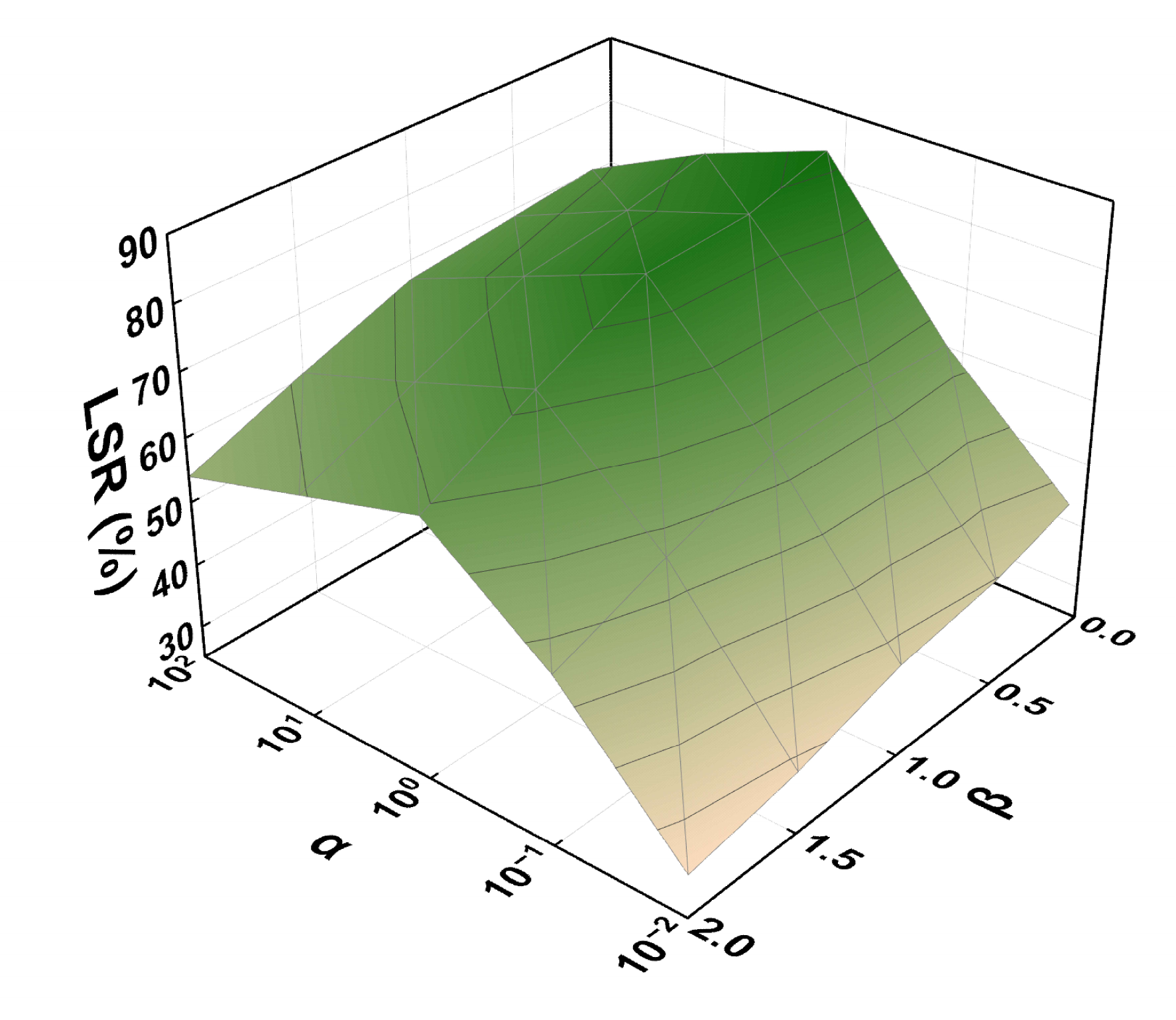}
        \caption{LSR (\%)}
        \label{Fig_Param_LSR}
    \end{subfigure}
    \hfill
    \begin{subfigure}[b]{0.49\columnwidth}
        \centering
        \includegraphics[width=\textwidth]{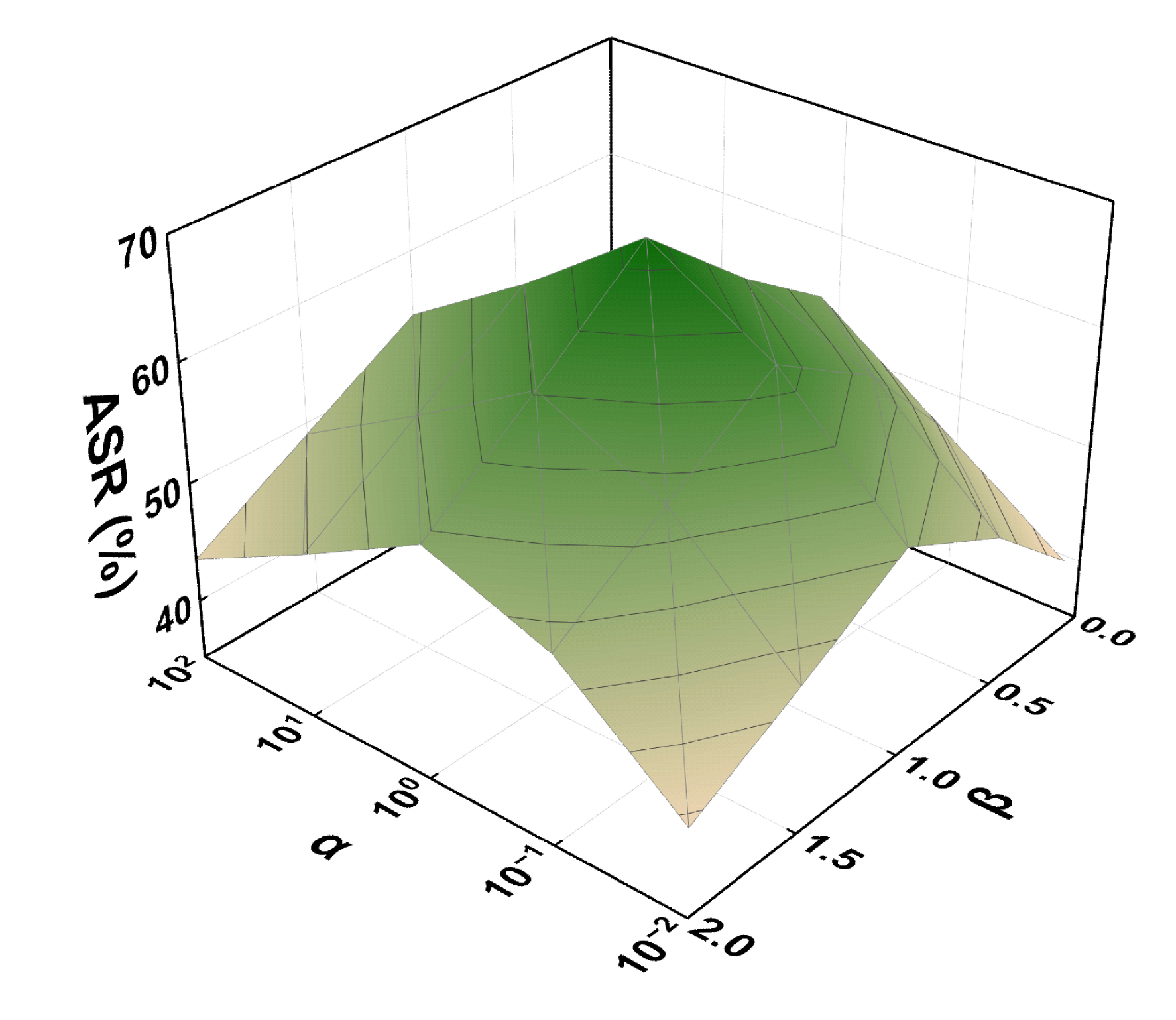}
        \caption{ASR (\%)}
        \label{Fig_Param_ASR}
    \end{subfigure}
    \caption{Hyperparameter sensitivity analysis.}
    \label{Fig_Param_Analysis}
\end{figure}

In this subsection, we delve deeper into how the hyperparameters $\alpha$ and $\beta$ influence the LSR and ASR of LiSA. As previously mentioned, $\alpha$ regulates feature optimization, while $\beta$ balances structure optimization. We examine their effects at a learning rate of $1.0$, testing $\alpha$ values of $\{0.01, 0.1, 1, 10, 100\}$ and $\beta$ values of $\{0.1, 0.5, 1, 1.5, 2\}$ on the PubMed dataset. As illustrated in Figure \ref{Fig_Param_Analysis}(a), LSR generally increases with increasing $\alpha$ and decreasing $\beta$. However, LSR decreases at higher $\alpha$ values because a larger $\alpha$ leads to bigger update steps at the same learning rate, hindering training convergence. As shown in Figure \ref{Fig_Param_Analysis}(b), higher $\alpha$ or lower $\beta$ reduces the adversarial effect of the subgraph, with an optimal ASR observed near $\alpha = 1$ and $\beta = 1$.

\section{Subgraph Training Time Analysis}

Table \ref{Table_Runtime} presents the subgraph training time of LiSA for each attack. The subgraphs of Cora/PubMed/FacebookPagePage, and Amazon-Photo/Computers were trained with 200 and 300 epochs, respectively, using an A6000 GPU with 48GB of memory. 

\begin{table}[htbp]
\centering
\caption{Subgraph training duration.}
\label{Table_Runtime}
\begin{tabular}{@{}lcccc@{}}
\Xhline{1.5pt}
\textbf{Dataset} & \textbf{Nodes} & \textbf{Edges} & \textbf{Epochs} & \textbf{Duration (sec)}\\
\Xhline{1.5pt}
\textbf{Cora} & 2,708 & 10,556 & 200 & 7.47 \\
\textbf{Photo} & 7,650 & 238,162 & 300 & 215.47 \\
\textbf{Computers} & 13,752 & 491,722 & 300 & 650.18 \\
\textbf{PubMed} & 19,717 & 88,648 & 200 & 625.38 \\
\textbf{PagePage} & 22,470 & 342,004 & 200 & 1,036.42 \\
\Xhline{1.5pt}
\end{tabular}
\end{table}

\end{document}